\documentclass[sigconf]{acmart}
\AtBeginDocument{%
  \providecommand\BibTeX{{%
    \normalfont B\kern-0.5em{\scshape i\kern-0.25em b}\kern-0.8em\TeX}}}

\usepackage{amsmath,graphicx}{}
\usepackage{bigstrut}
\usepackage{doi}
\usepackage{hyperref}
\usepackage{multirow,threeparttable, booktabs, makecell, pifont}
\usepackage{tabularx}

\usepackage{amssymb}
\usepackage{lipsum}

\usepackage{subfig}
\usepackage{float}

\begin{document}

\title{PeFoMed: Parameter Efficient Fine-tuning of Multimodal Large Language Models for Medical Imaging}

% \affiliation{%

\author{Jinlong He}
\authornotemark[1]
\email{hejinlong@hrbeu.edu.cn}
\affiliation{
	\institution{College of Computer Science and Technology}
	\streetaddress{Harbin Engineering University}
	\city{Harbin}
	\state{Heilongjiang}
	\country{China}
	\postcode{150001}}

\author{Pengfei Li}
\authornote{Both authors contributed equally to this research.}
\email{pengfeili@hrbeu.edu.cn}
\affiliation{
	\institution{College of Computer Science and Technology}
	\streetaddress{Harbin Engineering University}
	\city{Harbin}
	\state{Heilongjiang}
	\country{China}
	\postcode{150001}}

\author{Gang Liu}
\authornotemark[2]
\email{liugang@hrbeu.edu.cn}
\affiliation{
    \institution{College of Computer Science and Technology}
    \streetaddress{Harbin Engineering University}
    \city{Harbin}
    \state{Heilongjiang}
    \country{China}
    \postcode{150001}}

\author{Genrong He}
\email{hegenrong@hrbeu.edu.cn}
\affiliation{
    \institution{College of Computer Science and Technology}
    \streetaddress{Harbin Engineering University}
    \city{Harbin}
    \state{Heilongjiang}
    \country{China}
    \postcode{150001}}

\author{Zhaolin Chen}
\email{zhaolin.chen@monash.edu}
\affiliation{
    \institution{Monash Biomedical Imaging}
    \city{Melbourne}
    \state{Victoria}
    \streetaddress{Monash University}
    \country{Australia}
    \postcode{3800}}
\affiliation{
    \institution{Department of Data Science and AI, Faculty of Information Technology}
    \city{Melbourne}
    \state{Victoria}
    \country{Australia}
    \postcode{3800}}

\author{Shenjun Zhong}
\authornote{Corresponding author.}
\email{shenjun.zhong@monash.edu}
\affiliation{
    \institution{Monash Biomedical Imaging}
    \streetaddress{Monash University}
    \city{Melbourne}
    \state{Victoria}
    \country{Australia}
    \postcode{3800}}
\affiliation{
    \institution{National Imaging Facility}
    \city{Brisbane}
    \state{Queensland}
    \country{Australia}
    \postcode{4702}}
%% You do not have to enter your paper ID

\begin{abstract}
Multimodal large language models (MLLMs) represent an evolutionary expansion in the capabilities of traditional large language models, enabling them to tackle challenges that surpass the scope of purely text-based applications. It leverages the knowledge previously encoded within these language models, thereby enhancing their applicability and functionality in the reign of multimodal contexts. Recent works investigate the adaptation of MLLMs as a universal solution to address medical multi-modal problems as a generative task. In this paper, we propose a parameter efficient framework for fine-tuning MLLMs, specifically validated on medical visual question answering (Med-VQA) and medical report generation (MRG) tasks, using public benchmark datasets. We also introduce an evaluation metric using the 5-point Likert scale and its weighted average value to measure the quality of the generated reports for MRG tasks, where the scale ratings are labelled by both humans manually and the GPT-4 model. We further assess the consistency of performance metrics across traditional measures, GPT-4, and human ratings for both VQA and MRG tasks. The results indicate that semantic similarity assessments using GPT-4 align closely with human annotators and provide greater stability, yet they reveal a discrepancy when compared to conventional lexical similarity measurements. This questions the reliability of lexical similarity metrics for evaluating the performance of generative models in Med-VQA and report generation tasks. Besides, our fine-tuned model significantly outperforms GPT-4v. This indicates that without additional fine-tuning, multi-modal models like GPT-4v do not perform effectively on medical imaging tasks. The code will be available here: \url{https://github.com/jinlHe/PeFoMed}.
\end{abstract}






\maketitle

\section{Introduction}
\label{sec:intro}

Medical multimodal tasks involve the integration of both computer vision (CV) and natural language processing (NLP) techniques to analyze data from multiple modalities (i.e. image and text) to answer clinical-related questions as medical visual question answering (Med-VQA) tasks or generate textual reports from radiological images. Various deep learning models primarily approach medical multimodal tasks using dedicated models for each task, for example using classification models for VQA \cite{2,3,4,5,6,56} that categorize the image-text representation into a predefined set of answers, and auto-regressive models for generating image reports \cite{44,45,46}.

Recently, an alternative solution is treating medical multimodal tasks as generative tasks by using language models as decoders \cite{38}, which provides a universal framework to tackle medical multimodal problems. There has been emerging research that uses Large Language Models (LLMs) \cite{8,9,10,11} for generating free-form text as answers on Med-VQA tasks, \cite{48} using proper prompting techniques. This type of models, namely Multimodal Large Language Models (MLLMs) have been actively studied and the early attempts, such as Med-Flamingo \cite{34} and LLaVA-Med \cite{33} have shown the performance on various medical multimodal tasks.

However, directly training MLLMs from scratch for solving medical multimodal tasks requires numerous computational resources and large-scale annotated data. In response to these challenges, we propose a novel framework that uses Parameter-Efficient Fine-tuning (PEFT) techniques on MLLM foundation models for Med-VQA and medical report generation (MRG) tasks, namely the PeFoMed model. We utilize the pre-trained weights of a general domain LLM and ViT\cite{18}, which have been adeptly trained on diverse datasets, and fine-tune them with medical image-caption pairs and downstream Med-VQA and medical reports datasets. In training, the vision encoder and LLM are frozen, and only the vision projection layer and the low-rank adaptation layer (LoRA) \cite{22} are updated, which results in a minimal footprint of trainable parameters. Specific prompting templates are designed for the above fine-tuning process.

Furthermore, metrics based on lexical similarity, such as BLEU, often fall short of capturing semantic similarity between generated text and ground truth. In generative tasks like VQA and report generation, we utilize the GPT-4 model to assess the semantic similarity of generated answers or reports and compare it with human evaluations to examine consistency. For report generation, where the text can be lengthy, we employ a 5-point Likert scale to gauge the overall quality and coherence. This approach investigates the potential of GPT-4 as an accurate evaluation tool for large datasets, with carefully designed prompting templates.

The contributions of this work can be summarized as follows:

\begin{itemize}
\item[$\bullet$] We present a parameter-efficient method for fine-tuning general-domain foundation models for medical imaging applications, providing a universal solution for downstream tasks such as Med-VQA and MRG with minimal computational footprint. This approach challenges the specialised model design paradigm by utilizing LLMs as decoders.
\item[$\bullet$] We observed a discrepancy between accuracy measurements obtained through conventional lexical similarity and those assessed by semantic similarity using GPT-4 and human annotators. This discrepancy calls into question the reliability of lexical similarity metrics for evaluating the performance of generative models in Med-VQA and MRG tasks.
\item[$\bullet$] Our results show that semantic similarity assessments using GPT-4 demonstrate strong consistency with human annotators while offering enhanced stability.
\item[$\bullet$] Our fine-tuned model significantly outperforms GPT-4. This indicates that without additional fine-tuning, multi-modal models like GPT-4v do not perform effectively on medical imaging tasks.
\item[$\bullet$] We conducted experiments on public benchmarking datasets, including VQA-RAD, Slake, PathVQA and IU-Xray. The results show that our proposed model outperforms the latest generative models and achieves comparable performance to highly specialized models.
\end{itemize}

\section{Related works}
\subsection{Medical Multimodal Large Language Model}
\label{ssec:Medical multimodal large language model}
Large language foundation models, such as GPT-3 \cite{23}, PaLM \cite{24}, and LLaMA \cite{9} have demonstrated superior performance across a diverse range of medical-related NLP tasks. Previous works like ChatDoctor \cite{25}, Doctorglm \cite{26} and Huatuo \cite{27}, have yielded promising results in various medical NLP tasks. One step ahead, the latest works have leveraged both vision and language foundation models to resolve multimodal tasks in the medical domain aligns with this paradigm. Early attempts to address multimodal (i.e. vision and language) problems, like Visual Med-Alpaca \cite{32}, that converts images to intermediate text prompts, combine with the question texts and feed them into the LLM for predicting answers. This type of method may be constrained by the pre-trained image caption model and fail to capture detailed information from images.

Alternatively, recent work integrates vision embeddings into language models as visual prompts to enhance text generation. LLM-CXR \cite{40} applied a pre-trained VQ-GAN \cite{41} to tokenize images and generate visual and language tokens in an autoregressive manner, and fine-tuned the entire LLM with these tokens. Another intuitive solution is to explicitly model the projection of image embedding space to LLM space that can produce LLM-aligned image embeddings \cite{35}. Similarly, LLaVA-Med \cite{33} fine-tuned the projection layer and the entire LM on the GPT-4 generated instruction-tuning and downstream biomedical datasets.

Fine-tuning the entire LLM may not always be a practical solution when computing resources are constrained, instead, parameter-efficient fine-tuning techniques offer a balanced and efficient approach to adapt large pre-trained models to specific tasks while preserving the vast knowledge these models have already acquired. There have been some works that use PEFT techniques in various multimodality use cases \cite{14,42}. Therefore, in our approach, we employ parameter-efficient fine-tuning methods on Med-VQA and MRG tasks, which minimizes the training costs while yielding robust results.

\subsection{Medical Visual Question Answering}
\label{ssec:Medical Visual Question Answering}
Med-VQA tasks can be categorized into classification tasks \cite{2,3,4,5,6,7} and generative tasks \cite{33,34,35}. Classification-based Med-VQA predefines an answer candidate set. Although this approach can yield high performance on specific datasets, it concurrently limits the model’s capability to address open-ended questions. When integrated with LLMs, Med-VQA tasks transition to generative tasks. Li et al. explored the zero-shot setting of the GPT-4v model \cite{16} on Med-VQA dataset, where GPT-4v is a universal model and is not tailored to specific Med-VQA data types and tasks. Its performance on both open-ended and closed-ended question types is not comparable to the existing non-generative methods. Concurrently, Med-Flamingo \cite{34} extended from the base Flamingo framework \cite{39}, forming an in-context learning strategy with interleaved medical image-text pairs to achieve a few shot capacity for Med-VQA tasks. LLaVA-Med \cite{33} represents the first effort to adapt multimodal instruction tuning for the biomedical domain, implementing end-to-end training to develop biomedical multimodal dialogue assistants, thereby achieving promising outcomes in medical image-text dialogue.
\begin{figure*}[htb]
  \centering
  \centerline{\includegraphics[width=0.75\textwidth]{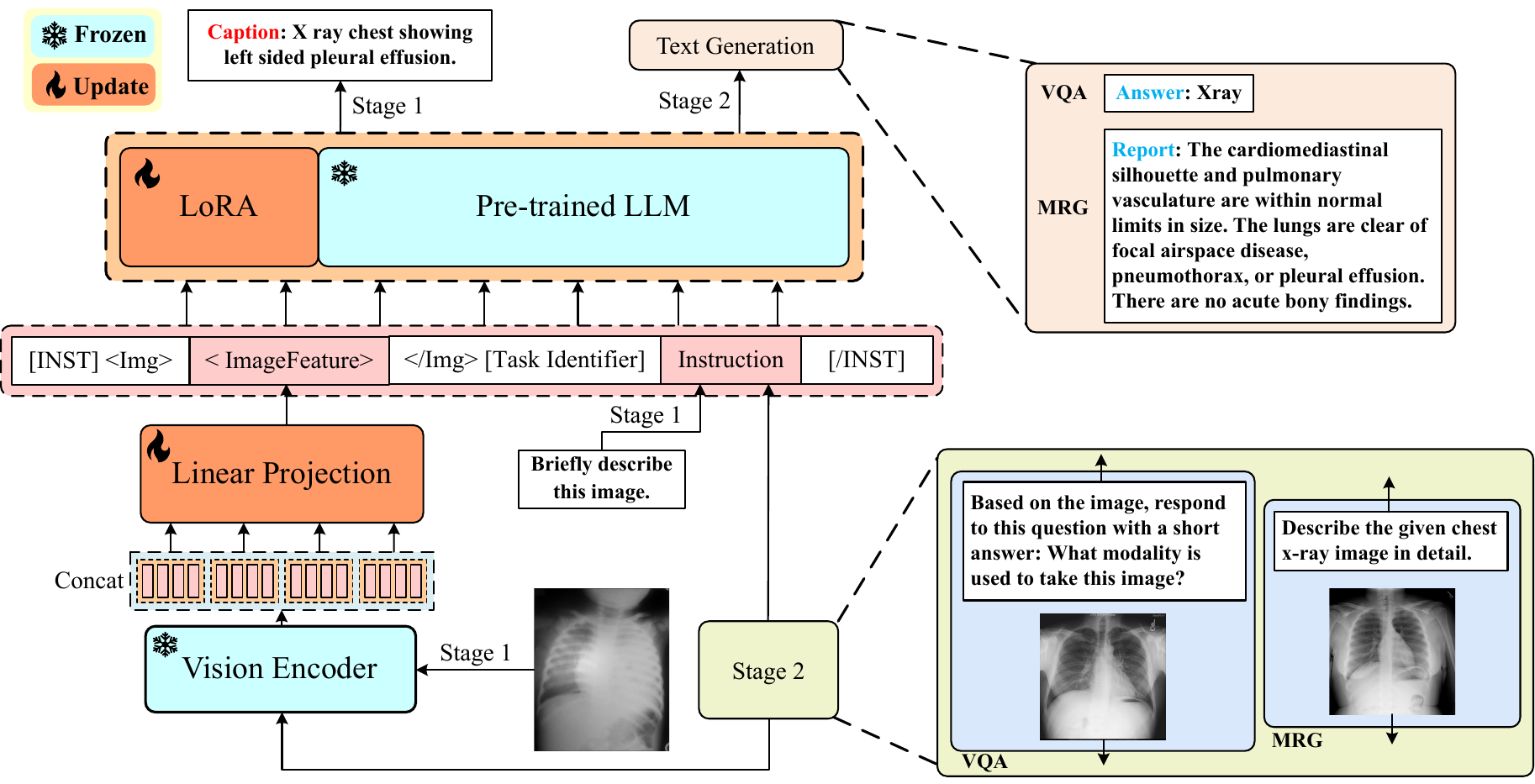}}
\caption{The architecture of the model.}
\label{fig:cdgcn}
\end{figure*}
\subsection{Medical Report Generation}
\label{ssec:Medical Report Generation}
Similar to image-captioning tasks, MRG generates text corresponding to medical images. As medical reports, the text generated by MRG is more professional than the conventional image-captioning task, and the text is relatively long and more complex. R2Gen \cite{47} produces radiology reports through a memory-driven transformer employing relational memory to capture key information throughout the generation process. Additionally, R2Gen employs conditional layer normalization to integrate memory within the transformer’s decoder. Chen et al. \cite{46} observed cross-modal data bias and applied cross-modal causal intervention to reduce this bias in medical reports and images, achieving notable results in medical report datasets. ITHN \cite{45} learns discriminative features between images and reports by separately learning images and their hard negatives, thereby capturing fine-grained details in multimodal data. Chen et al. \cite{44} addressed the discrepancy between medical images and text data by designing a distiller that leverages both prior and posterior knowledge to identify specific abnormalities in images and assimilate prior medical knowledge, yielding high performance in datasets such as IU-Xray.

Furthermore, Yang et al. introduced MedXChat \cite{49}, a model utilizing LLMs for chest X-ray multimodal tasks, which demonstrated effectiveness in generating chest X-ray medical reports. However, these studies often concentrate on a single MRG task or a specific type of medical image, like chest X-rays, employing general-domain text generation metrics for evaluation. Such metrics cannot accurately represent the quality of the generated medical reports.

\section{Methods}
\label{sec:Methods}
In this section, we will introduce the model architecture used in the work and the training strategy which involves two stages of fine-tuning. Subsequently, we will delineate the evaluation criteria adopted and the metrics proposed for MRG.

\subsection{Model Architecture}
\label{ssec:Model Architecture}

The model is composed of three parts: the vision encoder, a pre-trained LLM for processing multimodal inputs and generating answers, and a single linear layer for projecting embeddings from visual encoding space to LLM space, as shown in Fig.1. A ViT type of visual backbone, EVA \cite{18} is used to encode image tokens into visual embeddings, where the model weights are frozen during the entire fine-tuning processes. We group four consecutive tokens into one single visual embedding to effectively reduce resource consumption, by concatenating on the embedding dimension. The grouped visual tokens are then fed into the projection layer and output embeddings (of length 4096) in the LLM space. A multimodal prompt template is designed to include both visual and question information and passed to the pre-trained LLM, LLaMA2-chat(7B) \cite{11} as the decoder for generating answers. We adopt the low-rank adaptation (LoRA) technique \cite{22} in the LLM for efficient fine-tuning, where the other parts of the LLM are entirely frozen during the downstream fine-tuning. A beam search with a width of 1.

The multimodal prompt incorporates input images, questions and dedicated tokens for downstream tasks. In Figure 1, image features derived from linear projection are denoted as \textit{\textless ImageFeature\textgreater}, and the corresponding questions are the text instructions. Special tokens, such as \textit{[VQA]} for the Med-VQA task and \textit{[report]} for the MRG task, serve as task identifiers. This forms the complete multimodal instructional template as:\textit{[INST]\textless img\textgreater \textless ImageFeature\textgreater \textless /img\textgreater [Task Identifier] Instruction [/INST]}.

\subsection{Model Training}
\label{ssec:Model Training}
We use the weights from MiniGPT-v2 \cite{14} that are pre-trained on datasets in general domains, and further fine-tune the models using multimodal medical datasets in two stages. We adopt an efficient fine-tuning technique, LoRA \cite{22}, to only update a small part of the entire model. The details can be seen below:

\begin{sloppypar}
\textbf{Stage 1: Fine-tuning with Image Captioning.} During this stage, we fine-tune the model using four medical image-caption datasets, ROCO \cite{19}, CLEF2022 \cite{52}, MEDICAT \cite{51} and MIMIC-CXR \cite{50}. These datasets consist of medical image-caption, the captions are text descriptions of the corresponding images and have a variety of lengths. We use the prompt template: \textit{\textless Img\textgreater \textless ImageHere\textgreater \textless /Img\textgreater [caption] \textless instruction\textgreater}, and the instruction prompt used is randomly selected from a pool of four candidates, e.g. “Briefly describe this image”. During training, only the linear projection layer and the LoRA layer in the LLM are fine-tuned, while the other parts of the models are frozen.
\end{sloppypar}

\textbf{Stage 2: Fine-tuning on VQA and Report Generation.} In the second stage, the model undergoes fine-tuning on the Med-VQA and MRG datasets. Specifically, for Med-VQA, the VQA-RAD \cite{20}, SLAKE \cite{53}, and PathVQA \cite{54} datasets are used, while the IU-Xray \cite{55} dataset is utilized for the downstream MRG task. We adopt the following template, “\textit{[INST] \textless img\textgreater \textless ImageFeature \textgreater \textless /img\textgreater [Task Identifier] Instruction [/INST]}”, wherein \textit{[Task Identifier]} is substituted with \textit{[VQA]} or \textit{[Report]} according to the downstream tasks. For Med-VQA, the instruction prompt used in our experiment is: \textit{Based on the image, respond to this question with a short answer: \{question\}}, where \textit{\{question\}} is the question corresponding to the given medical image. The motivation for generating short answers is to validate against the annotated ground truth labels in VQA datasets where the answers are mostly short in both open-ended and closed-ended QA pairs. For MRG task, the instruction prompt is randomly selected from an instruction pool, for example: \textit{Describe the given chest x-ray image in detail}. Similarly, at this stage, we also keep the vision encoder and the LLM frozen while only updating the linear projection and LoRA layer in LLM.

\subsection{Evaluation Metrics}
\label{ssec:Evaluation Metrics}
In the experiment, we notice that the generated answers and the ground truth label in the VQA task, may not exactly match on a word-by-word basis, particularly for open-ended questions. As shown in Table 1, one of the common patterns from our observations is an inclusive relationship between the ground truth and the generated answer. For example, in the first case, it is shown where the ground truth is "x-ray" and the generated answer is "chest x-ray". In the conventional metric, this will be falsely classified as an incorrect prediction. In the second example, the prediction, "head" is semantically equivalent to the ground truth, "pancreatic head", where the organ information is presented in the question. Besides, there are cases where the prediction and ground truth are synonyms, like ‘both’ and ‘bilateral’ when asking questions about the sides of the lung.
%%%%%%%%%%%%%%%%%%%%%%%%%%%%%%%%%%%
\begin{table}[htbp]
  \centering
  \caption{Examples of model prediction on open-ended questions.}
  \resizebox{0.9\linewidth}{!}{
    \begin{tabular}{m{9.835em}cc}
    \toprule
    \textbf{Question} & \textbf{Ground Truth} & \textbf{Prediction} \\
    \midrule
    What kind of image is this? & x-ray      & chest x-ray \\
    \midrule
    The mass is found in which part of the pancreas? & pancreatic head & head \\
    \midrule
    Is the spleen present? & on patient's left & yes \\
    \midrule
    Are pleural opacities located on the left, right, or both sides of the lung? & both     & bilateral \\
    \bottomrule
    \end{tabular}}%
  \label{tab:addlabel}%
\end{table}%
%%%%%%%%%%%%%%%%%%%%%%%%%%%%%%%%%%%%%%%

This issue leads to an inaccurate measurement of model performance. Therefore, we evaluate all the generated answers from our model both manually and by GPT-4, against the ground truth in our experiment. We prepare a spreadsheet that contains all the predictions (that are labelled as not correct according to the conventional metric) with their associated ground truth. Then, ten team members independently evaluate each pair and re-classify the predictions as correct if they fall into the pattern as we shown above. In our experiment, we report both the exact match accuracy, GPT-4 measurement and the more precious measurement from our human evaluation process.

\begin{figure}[htb]
  \centering
  \centerline{\includegraphics[width=1\columnwidth]{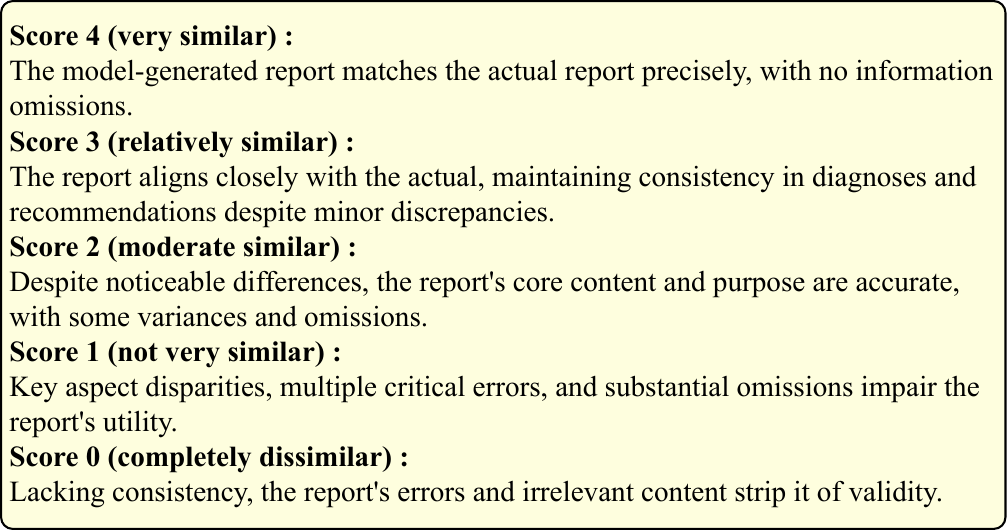}}
\caption{5-point Likert scale for semantic similarity, used to evaluate medical reports.}
\label{fig:cdgcn}
\end{figure}

\noindent
\textbf{Semantic Similarity Using GPT-4:} In the evaluation of generated results in open-ended questions of VQA and descriptive reports in MRG, we note that the existing evaluation metrics, based on lexical similarity, fall short in accurately capturing the semantic similarity measurements and handling synonyms. To mitigate this issue, we explore the options of using GPT-4 as the measurement engine to evaluate the semantic similarities between the ground truth text and its generated counterparts from the proposed models, which are then validated against human-labelled data.

%%%%%%%%%%%%%%%%%%%%%%%%%%%%%%%%%%%%%%%%% result table begin
\begin{table*}[htbp]
  \centering
  \caption{The performance of various models on the VQA-RAD, SLAKE, and PathVQA datasets, where numbers in brackets represent standard deviations, indicating that the results come from the average of ten independent evaluations.}
  \resizebox{1\textwidth}{!}{
    \begin{tabular}{c|c|c|c|ccc|ccc|ccc}
    \Xhline{1pt}
    \multirow{2}[4]{*}{\textbf{Methods}} & \multicolumn{1}{c|}{\multirow{2}[4]{*}{\textbf{Type}}} & \multicolumn{1}{c|}{\multirow{2}[4]{*}{\parbox{3cm}{\centering \textbf{Accuracy\\Measurement}}}} & \multicolumn{1}{c|}{\multirow{2}[4]{*}{\parbox{2cm}{\centering \textbf{Trainable\\Parameters}}}} & \multicolumn{3}{c|}{\textbf{VQA-RAD}} & \multicolumn{3}{c|}{\textbf{SLAKE}} & \multicolumn{3}{c}{\textbf{PathVQA}} \bigstrut\\
\cline{5-13}              &           &           &           & \textbf{Open} & \textbf{Closed} & \textbf{Overall} & \textbf{Open} & \textbf{Closed} & \textbf{Overall} & \textbf{Open} & \textbf{Closed} & \textbf{Overall} \bigstrut\\
    \Xhline{1pt}
    MMQ \cite{2}       & \multirow{7}[14]{*}{Non-LLMs} & \multicolumn{1}{c|}{\multirow{7}[14]{*}{Exact Match}} & 28.3M     & 52.0\%    & 72.4\%    & 64.3\%    & -         & -         & -         & 11.8\%    & 82.1\%    & 47.1\% \bigstrut\\
\cline{4-4}    MTL \cite{4}       &           &           & -         & 69.8\%    & 79.8\%    & 75.8\%    & 80.2\%    & 86.1\%    & 82.5\%    & -         & -         & - \bigstrut\\
\cline{4-4}    VQA-Adapter \cite{3} &           &           & 2.09M     & 66.1\%    & 82.3\%    & 75.8\%    & 79.2\%    & 83.7\%    & 81.0\%    & -         & -         & - \bigstrut\\
\cline{1-1}\cline{4-13}    M3AE \cite{5}      &           &           & -         & 67.2\%    & 83.5\%    & 77.0\%    & 80.3\%    & 87.8\%    & 83.2\%    & -         & -         & - \bigstrut\\
\cline{4-4}    M2I2 \cite{6}      &           &           & 262.15M   & 66.5\%    & 83.5\%    & 76.8\%    & 74.7\%    & 91.1\%    & 81.2\%    & 36.3\%    & 88.0\%    & 62.2\% \bigstrut\\
\cline{4-4}    MUMC \cite{7}      &           &           & 211.06M   & 71.5\%    & 84.2\%    & 79.2\%    & 81.5\%    & 91.1\%    & 84.9\%    & 39.0\%    & 90.4\%    & 65.1\% \bigstrut\\
\cline{4-4}    ARL \cite{36}       &           &           & 362M      & 67.6\%    & 86.8\%    & 79.2\%    & 81.9\%    & 91.4\% & 85.6\%    & -         & -         & - \bigstrut\\
    \hline
    LLaVA-Med(LLaVA) \cite{33} & \multicolumn{1}{c|}{\multirow{7}[10]{*}{LLMs}} & \multicolumn{1}{c|}{\multirow{3}[2]{*}{\parbox{3cm}{\centering Open: Token Recall \\ Closed: Exact Match}}} & \multirow{3}[2]{*}{7B} & 61.5\%    & 84.2\%    & 75.2\%    & 83.1\%    & 85.3\%    & 84.0\%    & 38.0\%    & 91.2\%    & 64.7\% \bigstrut[t]\\
    LLaVA-Med(Vicuna) \cite{33} &           &           &           & 64.4\%    & 82.0\%    & 75.0\%    & 84.7\%    & 83.2\%    & 84.1\%    & 38.9\%    & 91.7\% & 65.3\% \\
    LLaVA-Med(Bio-CLIP) \cite{33} &           &           &           & 64.8\%    & 83.1\%    & 75.8\%    & 87.1\% & 86.8\%    & 87.0\% & 39.6\%    & 91.1\%    & 65.4\% \bigstrut[b]\\
\cline{3-4}    GPT-4v \cite{16} &           & \multicolumn{1}{c|}{Exact Match} & -         & -         & 61.4\%   & -         & -         & -         & -         & -         & -         & - \bigstrut\\
\cline{1-1}\cline{3-13}    \multirow{3}[6]{*}{\textbf{PeFoMed (ours)}} &           & \multicolumn{1}{c|}{Exact Match} & \multirow{3}[6]{*}{56.63M} & 62.6\%    & 87.1\%    & 77.4\%    & 77.8\%    & 88.7\%    & 82.1\%    & 35.7\%    & 91.3\%    & 63.6\% \bigstrut\\
\cline{3-3}              &           & Human Evaluation &           & 77.7\%(3.2\%) & 87.6\%(0.2\%) & 83.7\%(1.3\%) & -         & -         & -         & -         & -         & - \bigstrut\\
\cline{3-3}              &           & GPT-4 Evaluation &           & 79.9\%(1.2\%) & 87.5\%(0.0\%) & 84.4\%(0.5\%) & 83.1\%(0.3\%) & 88.7\%(0.0\%) & 85.3\%(0.2\%) & 45.7\% & 91.3\%    & 68.6\% \bigstrut\\
    \Xhline{1pt}
    \end{tabular}}%
  \label{tab:addlabel}%
\end{table*}%
%%%%%%%%%%%%%%%%%%%%%%%%%%%%%%%%%%%%%%%%% result table end

Particularly in MRG tasks, where the generated content is relatively lengthy, using a binary rating may not be ideal. Instead, a 5-point Likert scale for semantic similarity can provide a more fine-grained validation of the results. As shown in Figure 2, it encompasses five levels: 0, 1, 2, 3, and 4. Each level represents specific evaluation criteria, with the quality of generated reports progressively improving from level 0 (completely dissimilar) to level 4 (very similar). We calculate the weighted average score of the similarity metric (WASM), as follows:
\begin{equation}
WASM = \frac { \sum _ { i = 0 } ^ { 4 } ( i \cdot N _ { i } ) } { \sum _ { i = 0 } ^ { 4 } N _ { i } }
\end{equation}
where $i$ denotes the ith level and $N _ { i }$ denotes the number of samples in the ith level. In our experiments, this metric is assessed through both human evaluation and GPT-4 evaluation.

\section{Experiments and results}
\label{sec:Experiments and results}

\subsection{Datasets}

We utilized four datasets for stage 1 image-caption fine-tuning: ROCO \cite{19}, CLEF2022 \cite{52}, MEDICAT \cite{51}, and MIMIC-CXR \cite{50}. ROCO, consisting of 87,952 radiological images and associated captions from PubMed Central, our study used only the training set. CLEF2022, offering broad coverage across medical fields, contains over 90,000 image-caption pairs. MEDICAT features over 210,000 image-caption pairs from PubMed Central. MIMIC-CXR features 377,110 radiology images and 227,835 reports from 64,588 patients. We used the official training set split.

Stage 2 fine-tuning for Med-VQA utilized three datasets: VQA-RAD \cite{20}, SLAKE \cite{53}, and PathVQA \cite{54}. VQA-RAD comprises 315 radiologic images with 3,515 question-answer pairs across 11 question types such as abnormality and modality, featuring 104 axial CT scans of the abdomen, 104 head scans (CT/MRI), and 107 chest radiographs, split into 3,064 training and 451 test pairs. SLAKE, a bilingual Med-VQA dataset, provided us with the English subset containing 642 images and 7,032 pairs, divided as 4,918 for training, 1,053 for validation, and 1,061 for testing. PathVQA features 4,998 pathology images and 32,795 question-answer pairs. It includes eight question types: 'what', 'where', 'when', 'how', 'why', 'whose', 'which', and 'how much'. The dataset was split into training, validation, and test sets in a 7:1:2 ratio.

The IU-Xray dataset was utilized for stage 2 fine-tuning in MRG. IU-Xray, a benchmark dataset from Indiana University, is widely used for MRG evaluation. It includes 7,470 chest X-ray images and 3,955 radiology reports, each associated with frontal and lateral views. The dataset was segmented into training, validation, and test sets in a 7:1:2 ratio for this study.

\subsection{Implementation Details}
All experiments were carried out using Python 3.9 on 4 NVIDIA Tesla A40 GPUs, each with 48GB of GPU memory. We initialized the model using the pre-trained weights of MiniGPT-v2. Throughout the training process, we only updated the linear projection layer and fine-tuned the LoRA layers (with a rank of 64) of the LLM.

In both fine-tuning stages, images are set to a resolution of $448\times 448$, and the maximum text length is set to 1024. We employed AdamW \cite{21} optimizer along with a cosine learning rate scheduler to train the model. For stage 1 fine-tuning, the learning rate was gradually decreased from an initial $1e^{-4}$ to $8e^{-5}$, the epoch is set to 3. In stage 2 fine-tuning, the learning rate is progressively lowered from $3e^{-5}$ to $1e^{-5}$, and the epoch is set to 50.

\subsection{Comparison with the State-of-the-Art Methods}
In this section, we present the results of our proposed models against the existing state-of-the-art (SOTA) models, on both Med-VQA and MRG tasks.

\noindent
\textbf{a. Medical Visual Question Answering}

For Med-VQA, as shown in Table 2, we compare the proposed model with the previous SOTA methods. We briefly categorized existing models based on their decoder types, including non-LLM types such as classifiers, BERT \cite{37}, and generative LLMs. For non-LLM approaches employing classifiers or BERT as decoders, the accuracy was computed through an exact match metric between the predicted answers and the ground truth text. On the other hand, the type of methods, like our method, used generative models that yield free-form text as answers using LLM as decoders, where the accuracy was measured differently. LLaVA-Med \cite{33} employed token-based recall for open-ended questions, while exact-match accuracy for closed-ended ones. In our work, we applied three accuracy metrics for the generated free-form answers: exact match accuracy as in the previous works, GPT-4 similarity evaluation and human manual evaluation in order to minimize the evaluation bias.

\begin{figure*}[htb]
  \centering
  \centerline{\includegraphics[width=0.85\textwidth]{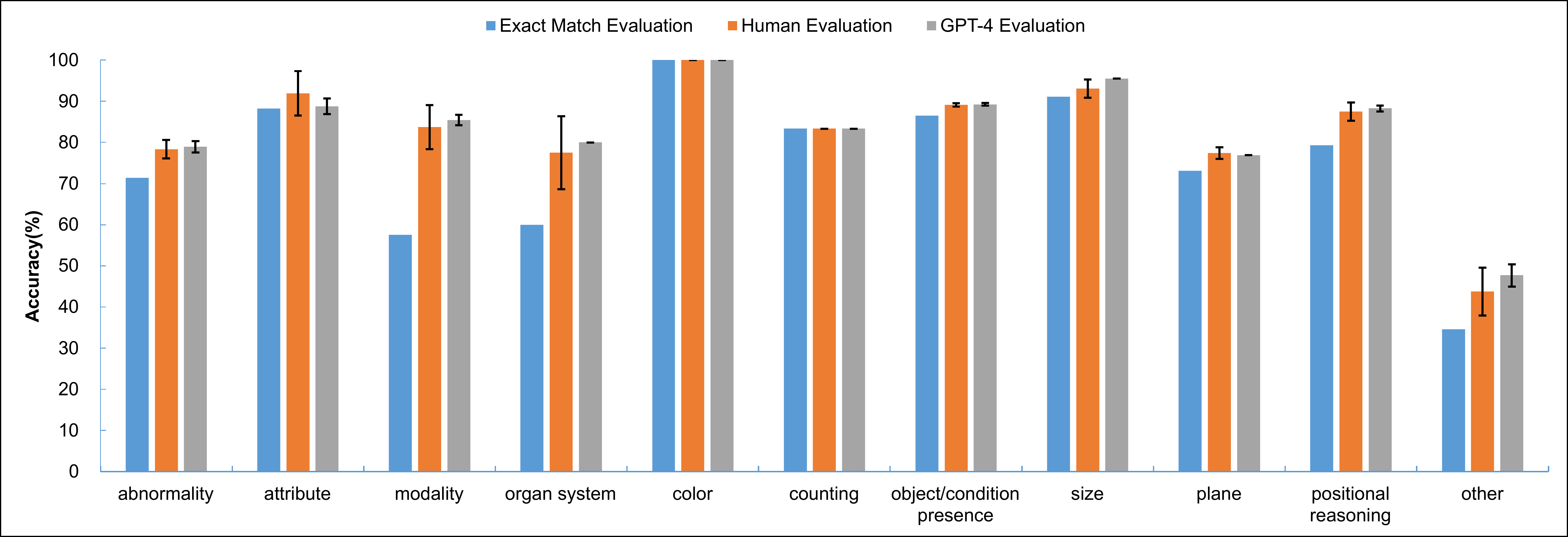}}
\caption{The accuracy of different question types on the VQA-RAD dataset by different evaluation methods.}
\label{fig:cdgcn}
\end{figure*}

On the VQA-RAD dataset, our LLM-based method was close to the performance of the dedicated VQA models, with an overall accuracy of 77.4\%, while outperforming the existing methods on closed-ended questions (87.1\%), under the same exact-match accuracy metric. For open-ended questions, our MLLM-based model achieved 62.6\%, compared to non-LLM dedicated visual language models, such as M3AE (67.2\%) \cite{5} and MUMC (71.5\%) \cite{7} In comparison, the early attempt to use LLM as answer decoders by the LLaVA-Med model series \cite{33} achieved overall accuracy of 75.8\%, specifically, 84.2\% for closed-ended questions and 64.8\% for open-ended questions. It is important to note that LLaVA-Med measures accuracy for open-ended and closed-ended questions differently, i.e. measuring token recall rate for open-ended questions, and conventional classification accuracy for closed-ended types. Furthermore, a recent study validated the capability of GPT-4v (without fine-tuning) on the VQA dataset, and reported an accuracy of 61.4\% on closed-ended questions, where our model achieved 87.1\% in comparison.

%%%%%%%%%%%%%%%%%%%%%%%%%%%%%%%%%%%%%%%%% result table
\begin{table}[htbp]
  \centering
  \caption{The accuracy of different phrase types on the VQA-RAD dataset by different evaluation methods, para denotes the parameterized form phrase.}
  \resizebox{0.7\linewidth}{!}{
    \begin{tabular}{c|cc}
    \hline
    \multicolumn{1}{c|}{\multirow{2}[4]{*}{\textbf{Evaluation Methods}}} & \multicolumn{2}{c}{\textbf{Phrase}} \bigstrut\\
\cline{2-3}    \multicolumn{1}{c|}{} & \textbf{freeform} & \textbf{para} \bigstrut\\
    \hline
    \textbf{Exact Match Evaluation} & 72.7\%    & 87.4\% \bigstrut[t]\\
    \textbf{Human Evaluation} & 81.5\%(1.8\%) & 88.2\%(0.4\%) \\
    \textbf{GPT-4 Evaluation} & 80.8\%(0.6\%) & 92.2\%(0.3\%) \bigstrut[b]\\
    \hline
    \end{tabular}}%
  \label{tab:addlabel}%
\end{table}%
%%%%%%%%%%%%%%%%%%%%%%%%%%%%%%%%%%%%%%%%% result table

As discussed in the method section, accuracy measured using the exact-match metric may not reflect the true performance in an MLLM-based generative setup. Therefore, we cross-validated the results using both GPT-4 and human evaluation methods, which were collected from 10 human annotators, and 10 independent inferences on the GPT-4 model with different seeding choices. Table 2 presents the results, indicating that the accuracies measured by human annotators (87.6\%) and GPT-4 (87.5\%) align closely with the exact match metric (87.1\%). However, a significant discrepancy is observed in open-ended questions, where using GPT-4 evaluation yields an accuracy of 79.9\%, closely mirroring the human annotations (77.7\%) and surpassing the exact matches. This discrepancy raises concerns about the suitability of the exact match metric for evaluating open-ended questions and MRG tasks.

We further analyzed the impacts of different evaluation methods across the pre-defined question types within the VQA-RAD dataset. Figure 3 shows the accuracy under the three distinct evaluation methods and the standard deviations for human and GPT-4 evaluations. The results suggest that accuracy measurements from human and GPT-4 evaluations are closely aligned across different question types, while GPT-4 has a consistently low variability compared to human annotators. A noticeable disparity can be seen between the results of human and GPT-4 evaluation methods and the exact match evaluation, for question types of 'abnormality', 'modality', and 'organ system'. Besides, regardless of the evaluation metrics, the model performed relatively well on question types, like ‘color’, ‘attribute’ and ‘size’. Questions associated with ‘abnormality’, ‘plane’ and ‘organ system’ seem to be challenging for our approach.

%%%%%%%%%%%%%%%%%%%%%%%%%%%%%%%%%%%%%%%%% result table
\begin{table}[htbp]
  \centering
  \caption{The accuracy of different image organs under different evaluation methods on the VQA-RAD dataset.}
  \resizebox{0.8\linewidth}{!}{
    \begin{tabular}{c|ccc}
    \hline
    \multicolumn{1}{c|}{\multirow{2}[4]{*}{\textbf{Evaluation Methods}}} & \multicolumn{3}{c}{\textbf{Organ}} \bigstrut\\
\cline{2-4}    \multicolumn{1}{c|}{} & \textbf{CHEST} & \textbf{HEAD} & \textbf{ABDOMEN} \bigstrut\\
    \hline
    \textbf{Exact Match Evaluation} & 81.0\%        & 79.0\%       & 72.2\%  \bigstrut[t]\\
    \textbf{Human Evaluation} & 86.4\%(1.0\%) & 85.4\%(2.8\%) & 79.4\%(1.9\%) \\
    \textbf{GPT-4 Evaluation} & 87.4\%(0.0\%)  & 86.8\%(1.3\%) & 79.3\%(0.7\%) \bigstrut[b]\\
    \hline
    \end{tabular}}%
  \label{tab:addlabel}%
\end{table}%
%%%%%%%%%%%%%%%%%%%%%%%%%%%%%%%%%%%%%%%%% result table

\begin{sloppypar}
A similar discrepancy can be seen on different phrase types (i.e. free-form and parameterized form) in the VQA-RAD dataset. The free-form question refers to a type of question or answer format that does not adhere to a predefined structure or set of possible responses, and are usually open-ended, allowing for a wide range of natural language expressions. Parameterized form phrase uses a structured representation of a question or answer that includes placeholders for specific parameters. For example, In the phrase, "What is the size of the [anatomical structure]?" where "[anatomical structure]" is a placeholder that can be replaced with different anatomical structures such as "heart", "lung", or "tumor" to generate specific questions like "What is the size of the heart?" As shown in Table 3, the discrepancy of exact-match and GPT-4 or human evaluation in free-form questions is much larger than the parameterized ones. As shown in Table 4, among the three organ types, the largest discrepancy between human and GPT-4 evaluations is merely 1.4\%, with the smallest being 0.1\%. The smallest gap with exact match evaluation reaches 5.4\%.
\end{sloppypar}

%%%%%%%%%%%%%%%%%%%%%%%%%%%%%%%%%%%%%%%%% result fig
\begin{figure}[htb]
  \centering
  \centerline{\includegraphics[width=0.8\columnwidth]{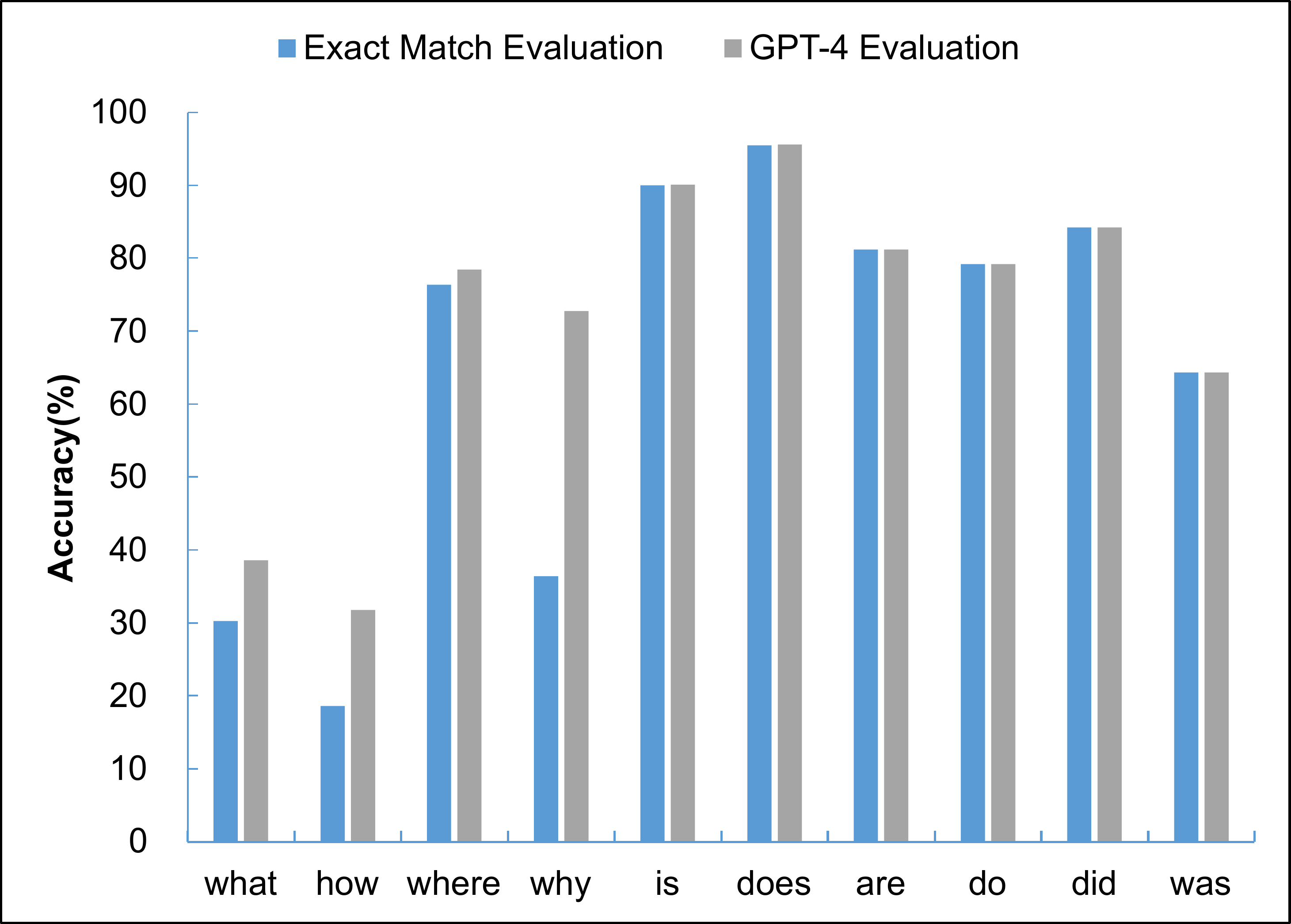}}
\caption{Accuracy of different evaluation methods for different types of questions on the PathVQA dataset.}
\label{fig:cdgcn}
\end{figure}
%%%%%%%%%%%%%%%%%%%%%%%%%%%%%%%%%%%%%%%%% result fig

Similarly, we observed significant performance gaps of accuracy measurements for open-ended questions on both the SLAKE and PathVQA datasets, using GPT-4 and exact-match evaluations. On the SLAKE dataset, our model achieved an overall accuracy of 82.1\% (with exact-match evaluation) and 85.3\% (using GPT-4 evaluation), showcasing a competitive performance across all evaluated models. On the PathVQA dataset, the largest among the three VQA datasets, featuring a test set of 6761 samples, there is an even larger discrepancy of 10\% for open-ended questions between the two evaluations.

Using GPT-4 semantic similarity evaluation, the results demonstrate a strong consistency across multiple inferences on the VQA-RAD and SLAKE datasets. The largest standard deviation across 10 independent runs is only 1.2\%, compared to the results of 10 independent human annotators (3.2\%) for open-ended questions. On closed-ended questions, the standard deviations on GPT-4 evaluation over multiple runs between generated answers and ground truth is nearly 0 on both VQA-RAD and SLAKE datasets.

%%%%%%%%%%%%%%%%%%%%%%%%%%%%%%%%%%%%%%%%% result table
\begin{table}[htbp]
  \centering
  \caption{Performance of various models on the IU-Xray.}
  \resizebox{0.8\linewidth}{!}{
    \begin{tabular}{c|c|ccc}
    \hline
    \multirow{2}[2]{*}{\textbf{Methods}} & \multicolumn{1}{c|}{\multirow{2}[2]{*}{\textbf{Type}}} & \multirow{2}[2]{*}{\textbf{METEOR$\uparrow$}} & \multirow{2}[2]{*}{\textbf{ROUGE-L$\uparrow$}} & \multirow{2}[2]{*}{\textbf{CIDEr$\uparrow$}} \bigstrut[t]\\
              &           &           &           &  \bigstrut[b]\\
    \hline
    PPKED \cite{44} & \multirow{4}[2]{*}{Non-LLMs} & \textbf{0.376} & 0.351     & - \bigstrut[t]\\
    ITHN \cite{45} &           & 0.210      & -         & \textbf{0.495} \\
    VLCL \cite{46} &           & 0.204     & \textbf{0.397} & 0.456 \\
    R2Gen \cite{47} &           & 0.211     & 0.377     & 0.438 \bigstrut[b]\\
    \hline
    BiomedGPT \cite{48} & \multicolumn{1}{c|}{\multirow{2}[2]{*}{LLMs}} & 0.146     & 0.302     & 0.360 \bigstrut[t]\\
    \textbf{PeFoMed (ours)} &           & 0.157     & 0.286     & 0.462 \bigstrut[b]\\
    \hline
    \end{tabular}}%
  \label{tab:addlabel}%
\end{table}%
%%%%%%%%%%%%%%%%%%%%%%%%%%%%%%%%%%%%%%%%% result table end

As illustrated in Figure 4, we provide a detailed analysis of the various types of questions contained in the PathVQA dataset. For most of the closed-ended question types in this dataset, like the questions of ‘is’, ‘does’ and ‘did’, the results using GPT-4 evaluation have almost identical results with the exact-match evaluation. On the other hand, observations revealed that for more abstract question types, including 'what', 'how', and 'why', the accuracy measurements using exact match evaluation significantly differ from the counterparts using GPT-4 semantic similarity evaluation. This indicates that exact match evaluation may not be sufficiently effective in assessing the performance of MLLMs on more complex open-ended question types.

Besides, it is worth highlighting that with the adoption of parameter-efficient fine-tuning, our model featured notably fewer trainable parameters (56.63M) in comparison to LLaVA-Med (7B parameters), which made our method a significantly more efficient framework for fine-tuning MLLMs.

\noindent
\textbf{b. Medical Report Generation}

As shown in Table 6, our MLLM-based model was evaluated on MRG tasks using the IU-Xray dataset. We evaluated the performance with the commonly used metrics: METEOR, ROUGE-L and CIDEr. The results show that the LLM-based models, including both our method and a recent approach, BiomedGPT \cite{48} underperformed the existing dedicated MRG models, while our method is slightly better than BiomedGPT on METEOR and CIDEr measurements.

%%%%%%%%%%%%%%%%%%%%%%%%%%%%%%%%%%%%%%%%% result table
\begin{table}[htbp]
  \centering
  \caption{Semantic similarity for 100 randomly selected samples from the IU-Xray dataset, assessed through both human evaluation and GPT-4 evaluation, respectively.}
  \resizebox{1\linewidth}{!}{
    \begin{tabular}{c|cccccc}
    \hline
    \textbf{Evaluation Methods} & \multicolumn{1}{c}{\textbf{0}} & \textbf{1} & \textbf{2} & \textbf{3} & \textbf{4} & \boldmath{}\textbf{WASM$\uparrow$}\unboldmath{} \bigstrut\\
    \hline
    \textbf{Human Evaluatioin} & \multicolumn{1}{c}{7.75(5.55)} & 23.38(12.81) & 21.88(5.64) & 29.13(14.73) & 17.25(11.07) & 2.24(0.42) \bigstrut[t]\\
    \textbf{GPT-4 Evaluation} & 2.80(0.63) & 7.80(1.62) & 17.10(1.91) & 22.60(5.36) & 48.70(2.54) & 3.08(0.03) \bigstrut[b]\\
    \hline
    \end{tabular}}%
  \label{tab:addlabel}%
\end{table}%
%%%%%%%%%%%%%%%%%%%%%%%%%%%%%%%%%%%%%%%%% result table end

Evaluating the WASM using GPT-4 on the IU-Xray dataset (covering 1180 samples) to measure the similarity between generated reports and ground truth, our model achieved a WASM score of 2.79, as shown in Table 7. Nearly 90\% of the similarity scores are within the range from 2 (moderate similar) to 4 (very similar), and 27.5\% of the generated reports were rated as highly similar to the ground truth. Furthermore, we randomly sampled 100 samples from the IU-Xray dataset and compared the score of the 5-point Likert scale between the GPT-4 and human evaluations, as shown in Table 6. An interesting observation from the study is that human annotators tend to have more neutral opinions; approximately 75\% of their responses fall within the 2-3 range on a 5-point scale. However, GPT-4 frequently assigned a higher similarity score of 4 in 48.7\% of cases. Using GPT-4 evaluation results in a significantly higher WASM score of 3.08, with a standard deviation of 0.03 across 10 runs, compared to a score of 2.24 when evaluated by humans. This raises the question of whether using GPT-4 for similar measurements provides a more robust metric for MRG tasks.

%%%%%%%%%%%%%%%%%%%%%%%%%%%%%%%%%%%%%%%%% result table
\begin{table}[htbp]
  \centering
  \caption{Semantic similarity evaluated by GPT-4 on the IU-Xray test set. Num denotes the count of samples at each level. Per indicates the percentage of samples corresponding to each level within the test set.}
  \resizebox{0.65\linewidth}{!}{
    \begin{tabular}{c|ccccc}
    \hline
    \multicolumn{1}{c|}{} & \textbf{0} & \textbf{1} & \textbf{2} & \textbf{3} & \textbf{4} \bigstrut\\
    \hline
    \multicolumn{1}{c|}{\textbf{Num}} & 22        & 118       & 267       & 449       & 324 \bigstrut[t]\\
    \multicolumn{1}{c|}{\textbf{\%}} & 1.9\%     & 10.0\%    & 22.6\%    & 38.1\%    & 27.5\% \bigstrut[b]\\
    \hline
    \boldmath{}\textbf{WASM$\uparrow$}\unboldmath{} & \multicolumn{5}{c}{2.79} \bigstrut\\
    \hline
    \end{tabular}}%
  \label{tab:addlabel}%
\end{table}%
%%%%%%%%%%%%%%%%%%%%%%%%%%%%%%%%%%%%%%%%% result table end

\subsection{Ablation Study}
%%%%%%%%%%%%%%%%%%%%%%%%%%%%%%%%%%%%%%%%% result table
\begin{table*}[htbp]
  \centering
  \caption{The ablation study under different training setups and the comparison between exact match evaluation and GPT-4 evaluation on the Slake and PathVQA datasets.}
  \resizebox{0.7\linewidth}{!}{
    \begin{tabular}{c|cc|ccc|ccc}
    \Xhline{1pt}
    \multirow{2}[3]{*}{\textbf{Evaluation Methods}} & \multirow{2}[3]{*}{\textbf{Stage1}} & \multirow{2}[3]{*}{\textbf{Stage2}} & \multicolumn{3}{c|}{\textbf{Slake}} & \multicolumn{3}{c}{\textbf{PathVQA}} \bigstrut[b]\\
\cline{4-9}              &           &           & \textbf{Open} & \textbf{Closed} & \textbf{Overall} & \textbf{Open} & \textbf{Closed} & \textbf{Overall} \bigstrut\\
    \Xhline{1pt}
    \multicolumn{1}{c|}{\multirow{4}[2]{*}{\parbox{2cm}{\centering\textbf{(a)\\Exact Match\\Evaluation}}}} & ×         & ×         & 23.9\%    & 58.9\%    & 37.6\%    & 2.4\%     & 59.0\%    & 30.8\% \bigstrut[t]\\
              & \checkmark & ×         & 11.6\%    & 50.5\%    & 26.9\%    & 1.2\%     & 53.3\%    & 27.3\% \\
              & ×         & \checkmark & 77.5\%    & 85.3\%    & 80.6\%    & 31.2\%    & 90.0\%    & 60.7\% \\
              & \checkmark & \checkmark & \textbf{77.8\%}    & \textbf{88.7\%}    & \textbf{82.1\%}    & \textbf{35.7\%}    & \textbf{91.3\%}    & \textbf{63.6\%} \bigstrut[b]\\
    \hline
    \multicolumn{1}{c|}{\multirow{4}[2]{*}{\parbox{2cm}{\centering\textbf{(b)\\GPT-4\\Evaluation}}}} & ×         & ×         & 47.0\%(0.5\%) & 59.9\%(0.0\%) & 52.1\%(0.3\%) & 12.3\%    & 59.3\%    & 35.9\% \bigstrut[t]\\
              & \checkmark & ×         & 50.0\%(0.9\%) & 62.5\%(0.4\%) & 54.9\%(0.7\%) & 12.1\%    & 53.9\%    & 33.1\% \\
              & ×         & \checkmark & 82.8\%(0.3\%) & 85.3\%(0.0\%) & 83.8\%(0.2\%) & 38.5\%    & 90.0\%    & 64.3\% \\
              & \checkmark & \checkmark & \textbf{83.1\%}(0.3\%) & \textbf{88.7\%}(0.0\%) & \textbf{85.3\%}(0.2\%) & \textbf{45.7\%}   & \textbf{91.3\%}    & \textbf{68.6\%} \bigstrut[b]\\
    \Xhline{1pt}
    \end{tabular}}%
  \label{tab:addlabel}%
\end{table*}%
%%%%%%%%%%%%%%%%%%%%%%%%%%%%%%%%%%%%%%%%% result table
\textbf{a. Medical Visual Question Answering}

We conducted an ablation study to explore the impacts of the two stage fine-tuning strategy on the VQA performance. In Table 8 and 9, they show the VQA accuracy measurements of models that (i) perform zero-shot without fine-tuning; (ii) are only fine-tuned on the image caption dataset; (iii) are only fine-tuned on the VQA dataset; and (iv) have two stage fine-tuning.

%%%%%%%%%%%%%%%%%%%%%%%%%%%%%%%%%%%%%%%%% result table
\begin{table}[htbp]
  \centering
  \caption{The ablation study under different training setups and the comparison between exact match evaluation, human evaluation and GPT-4 evaluation on the VQA-RAD dataset.}
  \resizebox{0.9\linewidth}{!}{
    \begin{tabular}{c|cc|ccc}
    \Xhline{1pt}
    \multicolumn{1}{c|}{\textbf{Evaluation Methods}} & \textbf{Stage1} & \textbf{Stage2} & \textbf{Open} & \textbf{Closed} & \textbf{Overall} \bigstrut[t]\\
    \Xhline{1pt}
    \multirow{4}[1]{*}{\parbox{2cm}{\centering\textbf{(a)\\Exact Match\\Evaluation}}} & ×         & ×         & 13.4\%    & 48.2\%    & 34.4\% \\
    \multicolumn{1}{c|}{} & \checkmark & ×         & 16.2\%    & 59.9\%    & 42.6\% \\
    \multicolumn{1}{c|}{} & ×         & \checkmark & 58.1\%    & 82.0\%    & 72.5\% \\
    \multicolumn{1}{c|}{} & \checkmark & \checkmark & \textbf{62.6\%}    & \textbf{87.1\%}    & \textbf{77.4\%} \bigstrut[b]\\
    \hline
    \multirow{4}[2]{*}{\parbox{2cm}{\centering\textbf{(b)\\GPT-4\\Evaluation}}} & ×         & ×         & 26.2\%(1.1\%) & 49.1\%(0.3\%) & 40.0\%(0.4\%) \bigstrut[t]\\
    \multicolumn{1}{c|}{} & \checkmark & ×         & 37.3\%(1.1\%) & 60.2\%(0.1\%) & 51.1\%(0.5\%) \\
    \multicolumn{1}{c|}{} & ×         & \checkmark & 75.0\%(0.4\%) & 82.7\%(0.0\%) & 79.6\%(0.1\%) \\
    \multicolumn{1}{c|}{} & \checkmark & \checkmark & \textbf{79.9\%(1.2\%)} & \textbf{87.5\%}(0.0\%) & \textbf{84.4\%(0.5\%)} \bigstrut[b]\\
    \hline
    \textbf{(c) Human Evaluation} & \checkmark & \checkmark & \textbf{77.7\%}(3.2\%) & \textbf{87.6\%(0.2\%)} & \textbf{83.7\%}(1.3\%) \bigstrut\\
    \hline
    \end{tabular}}%
  \label{tab:addlabel}%
\end{table}%
%%%%%%%%%%%%%%%%%%%%%%%%%%%%%%%%%%%%%%%%% result table

Tables 8 and 9(b) demonstrate that two-stage fine-tuning significantly enhances model performance across all three benchmark datasets in both exact-match and GPT-4 similarity evaluations, and second-stage fine-tuning contributes the most performance improvements. We primarily focus on the results from GPT-4 similarity evaluations, which are largely consistent with those from exact-match evaluations. Without fine-tuning the VQA-RAD dataset, models only achieve a 40\% overall accuracy and 26.2\% on open-ended questions. Stage 1 fine-tuning with image captioning tasks raises overall accuracy by 11.1\%, while Stage 2 fine-tuning on VQA tasks alone doubles performance to 79.6\% and boosts combined overall accuracy to 84.4\%. Similarly, without fine-tuning, the Slake dataset starts with open-ended question accuracy at 47.0\% and overall accuracy at 52.1\%, with modest gains from image-caption task fine-tuning. Direct VQA task fine-tuning substantially improves Slake's and PathVQA's accuracies to 83.8\% and 64.3\%, respectively, with overall accuracies reaching 85.3\% for Slake and 68.6\% for PathVQA after completing both stages, notably increasing PathVQA's open-ended question accuracy by 7.2\%.

Performance declines were observed when applying stage-1 fine-tuning alone, which involves training with image captioning tasks. This decline was particularly noticeable in the Slake and PathVQA datasets (in Table 8), where stage-1 fine-tuning significantly reduced performance on both open-ended and closed-ended questions. Specifically, applying only stage-1 fine-tuning leads to a significant decrease in accuracy, with a 10.7\% overall drop and a 12.3\% reduction in open-ended question accuracy on the Slake dataset. Combining image captioning tasks with VQA tasks leads to optimal performance, outperforming the use of either task alone.

\begin{table}[htbp]
  \centering
  \caption{The ablation study under different training setups on the IU-Xray dataset.}
  \resizebox{0.7\linewidth}{!}{
    \begin{tabular}{cc|ccc}
    \hline
    \textbf{Stage1} & \textbf{Stage2} & \textbf{METEOR$\uparrow$} & \textbf{ROUGE-L$\uparrow$} & \textbf{CIDEr$\uparrow$} \bigstrut\\
    \hline
    ×         & ×         & \multicolumn{1}{c}{0.059} & \multicolumn{1}{c}{0.089} & 0.007 \bigstrut\\
    \checkmark & ×         & \multicolumn{1}{c}{0.153} & \multicolumn{1}{c}{0.247} & 0.124 \bigstrut\\
    ×         & \checkmark & \multicolumn{1}{c}{0.146} & \multicolumn{1}{c}{0.295} & 0.393 \bigstrut\\
    \checkmark & \checkmark & 0.157     & 0.286     & 0.462 \bigstrut\\
    \hline
    \end{tabular}}%
  \label{tab:addlabel}%
\end{table}%

We observed discrepancies between the exact match and GPT-4 evaluations in the Slake dataset. Table 8 shows that after stage-1 fine-tuning, overall accuracy declined from 37.6\% to 26.9\%. However, using GPT-4 evaluation, open-ended and closed-ended accuracies showed an approximate increase of 3\%. This raises concerns about the reliability of using the exact match similarity metric (where accuracy is determined by a word-by-word comparison between predicted answers and ground truths) to evaluate the performance of generative models in Med-VQA. Such a metric can result in biased and inaccurate evaluations for models that used generative LLMs as answer decoders, particularly for open-ended questions.

\noindent
\textbf{b. Medical Report Generation}

Since the MRG task is akin to the image captioning task used in stage-1 fine-tuning, applying only stage-1 fine-tuning can yield significant performance improvements. Table 10 shows that zero-shot testing on the IU-Xray dataset produced poor results, with METEOR, ROUGE-L, and CIDEr scores of 0.059, 0.089, and 0.007, respectively. This is primarily because models trained in general domains may not perform well on medical images. After fine-tuning with image captioning, these metrics improved to 0.152 for METEOR, 0.089 for ROUGE-L, and 0.124 for CIDEr.  The model exhibited its best performance after undergoing both stages of fine-tuning, with the metrics reaching 0.157, 0.286, and 0.462.

\begin{table}[htbp]
  \centering
  \caption{The ablation study reported on semantic similarity by GPT-4 under different training setups on the IU-Xray dataset.}
  \resizebox{0.8\linewidth}{!}{
    \begin{tabular}{cc|cccccc}
    \hline
    \textbf{Stage1} & \textbf{Stage2} & \textbf{0} & \textbf{1} & \textbf{2} & \textbf{3}         & \textbf{4} & \textbf{WASM$\uparrow$} \bigstrut\\
    \hline
    ×         & ×         & 1048      & 79        & 33        & 20        & 0 & 0.17 \bigstrut[t]\\
    \checkmark & ×         & 51        & 170       & 192       & 318       & 449       & 2.80 \\
    ×         & \checkmark & 39        & 125       & 193       & 447       & 376       & 2.84 \\
    \checkmark & \checkmark & 22        & 118       & 267       & 449       & 324       & 2.79 \bigstrut[b]\\
    \hline
    \end{tabular}}%
  \label{tab:addlabel}%
\end{table}%

Additionally, we discovered that the Weighted Average Score Metric (WASM) is sensitive to changes in accuracy measurements. Table 11 shows the WASM scores for each experiment setting, where the model's WASM score without any fine-tuning stands at merely 0.17, with a total of 1127 reports receiving scores of 0 and 1. Following the first and second stage of medical image-caption fine-tuning, the WASM score improved to 2.8 and 2.84 respectively. This is consistent with the overall model performance as shown in Table 10. Moreover, fine-tuning models on tasks that resemble the downstream tasks seems to provide the greatest benefits. For instance, fine-tuning on image captioning, which is closely related to report generation, tends to enhance performance directly on MRG tasks.

\section{Conclusion}
\label{sec:conclusion}
In this work, we proposed a novel parameter efficient fine-tuning framework for fine-tuning multimodal large language models for both Med-VQA and MRG tasks using a generative approach. Our research challenges the specialized models by introducing an LLM-based decoder as a universal solution for diverse downstream tasks, with minimal training footprint. Additionally, we express concerns that current metrics based on lexical similarity may produce inaccurate results. As an alternative, we suggest using GPT-4 to assess the semantic similarity between generated content and the corresponding ground truths. To validate that, for the first time in the context of medical MLLMs, we employed both human and GPT-4 evaluations alongside conventional metrics for Med-VQA and MRG tasks. The findings indicate that GPT-4 is a robust and reliable semantic similarity metric, consistent with human evaluations but with greater stability, which makes it suitable for analyzing the quality of generated outcomes in Med-VQA and MRG tasks.

\section{Acknowledgments}
\label{sec:Acknowledgments}
This work was supported by the Ministry of Education Humanities and Social Science Research Planning Fund Project under grant number 23YJAZH079 and Natural Science Foundation of Heilongjiang Province under grant number LH2021F015.
% \begin{acks}
% \end{acks}

\bibliographystyle{ACM-Reference-Format}

\appendix

\section{More Qualitative Results}

We conducted detailed experiments on the Slake dataset, as illustrated in Figure 5. Experiments were carried out on various question types within the dataset. Notably, the "KG" type questions, those requiring additional knowledge for answers, exhibited the largest discrepancy post-evaluations. Possessing substantial additional knowledge precisely characterizes medical MLLMs. As indicated in Figure 6, experiments across various image organ types revealed that "Chest mediastinal" was the sole organ type yielding consistent results. This organ type is notably prevalent in the MIMIC-CXR dataset, which possesses the most extensive data used for first-stage image-caption fine-tuning. This observation further demonstrates the significant impact of stage 1 fine-tuning on downstream tasks.
\begin{figure}[htb]
  \centering
  \centerline{\includegraphics[width=1\columnwidth]{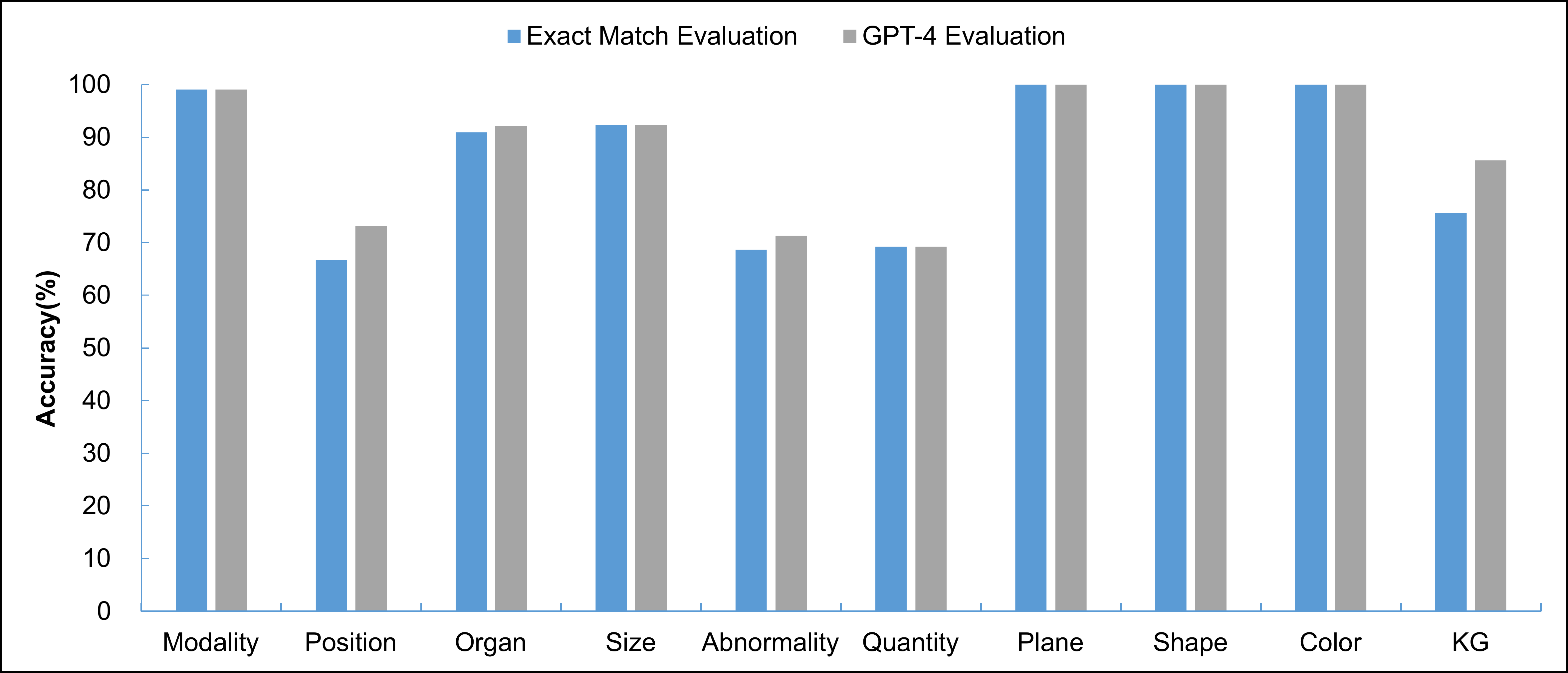}}
\caption{The accuracy of different evaluation methods for different question types on the Slake dataset.}
\label{fig:cdgcn}
\end{figure}

\begin{figure}[htb]
  \centering
  \centerline{\includegraphics[width=1\columnwidth]{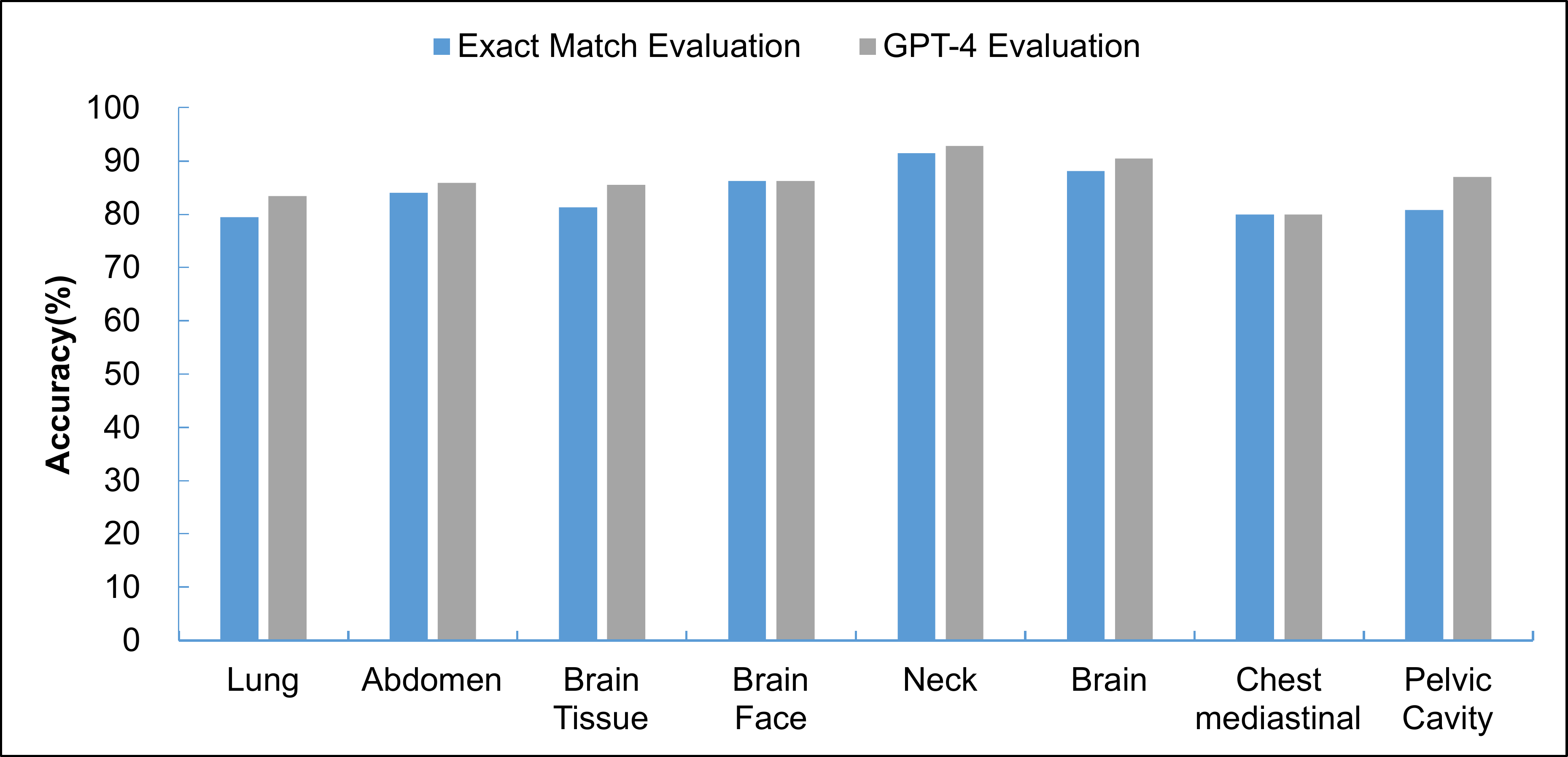}}
\caption{The accuracy of different evaluation methods for different image organ types on the Slake dataset.}
\label{fig:cdgcn}
\end{figure}

\begin{figure}[htb]
  \centering
  \centerline{\includegraphics[width=1\columnwidth]{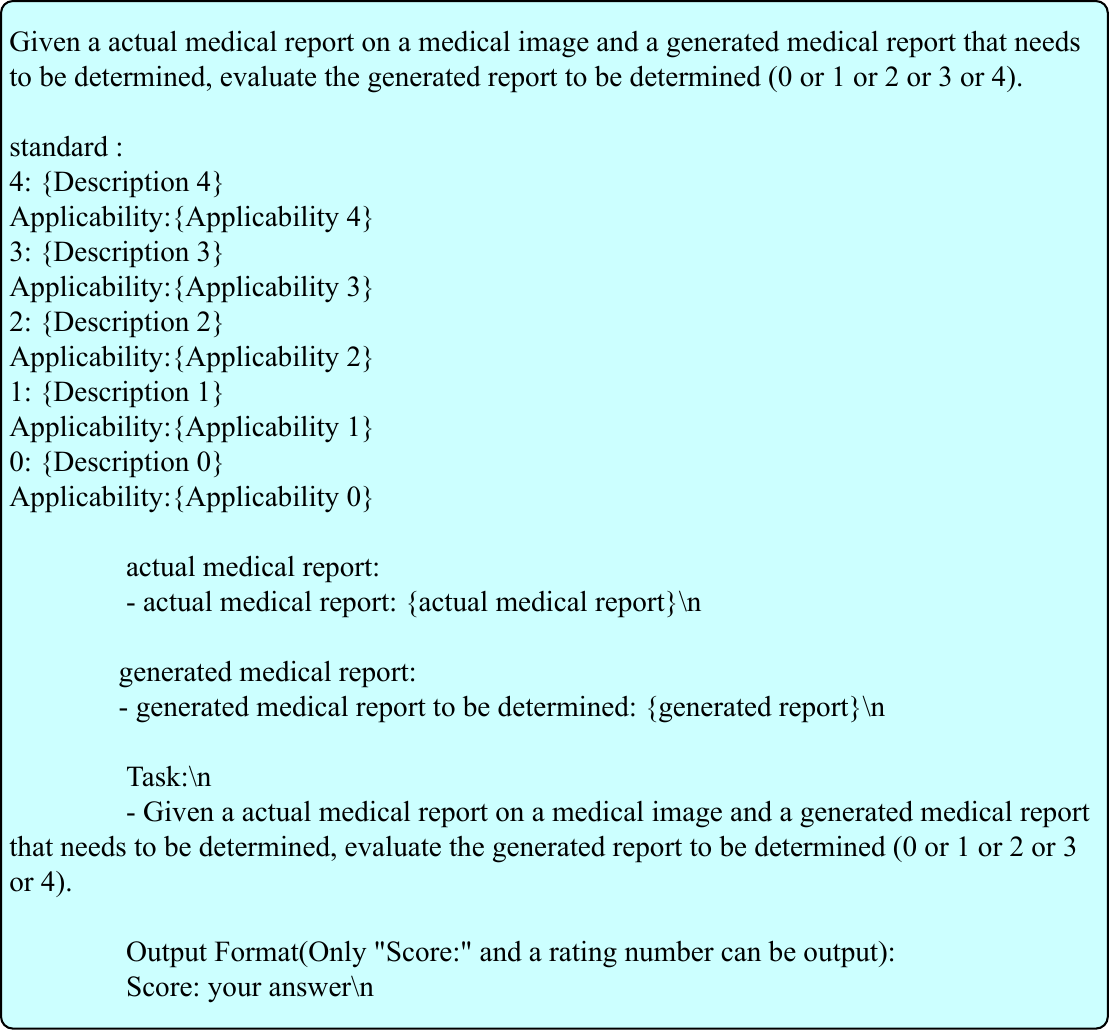}}
\caption{Instruction designed for GPT-4 to evaluate MRG.}
\label{fig:cdgcn}
\end{figure}

\begin{figure}[htb]
  \centering
  \centerline{\includegraphics[width=1\columnwidth]{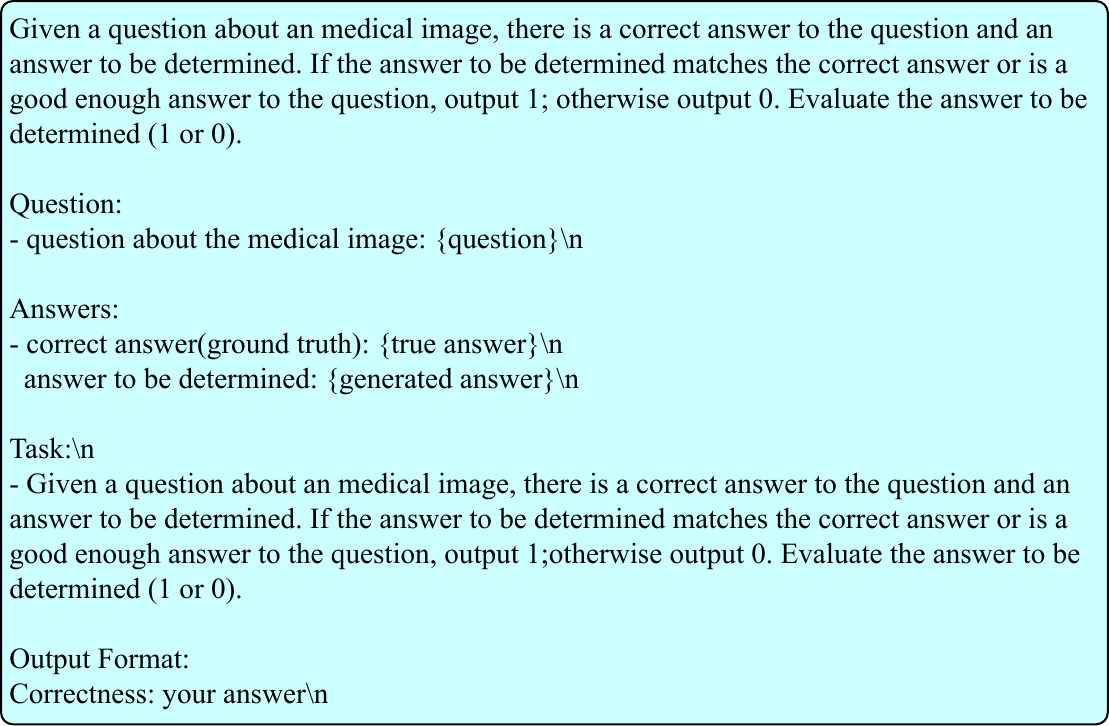}}
\caption{Instruction designed for GPT-4 to evaluate Med-VQA.}
\label{fig:cdgcn}
\end{figure}

\section{GPT-4 Instruction}

\subsection{MRG Instruction}
As shown in Figure 7, we formulated the instruction for GPT-4 evaluation in MRG, where {Description i} corresponds to the description for score $i$, {Applicability i} denotes the applicability of the score $i$, {actual medical report} refers the ground truth of the dataset, and {generated report} denotes the report produced by the model.

\subsection{Med-VQA Instruction}
The GPT-4 instruction designed by us to evaluate Med-VQA is shown in Figure 8, where {question} represents the question text in the VQA dataset, {true answer} represents the ground truth in the dataset, {generated answer} represents the answer generated by the model.

\section{More MRC Details}
As shown in Table 12, we give a detailed description and application of MRC metric, which is adopted by human evaluation and GPT-4 evaluation.

\begin{table*}[htbp]
  \centering
  \caption{Detailed description and applicability of MRC metric.}
    \begin{tabular}{c|m{8cm}|m{7cm}}
    \Xhline{1pt}
    \textbf{Score} & \textbf{Description} & \textbf{Applicability} \\
    \Xhline{1pt}
    \textbf{4} & The generated medical report exhibits minimal discrepancy in meaning compared to the actual report, ensuring complete consistency in critical aspects such as diagnosis and treatment recommendations. It encapsulates all requisite information elements without omissions. & This level is pertinent when evaluators face difficulty distinguishing between the generated and actual medical reports. \bigstrut\\
    \hline
    \textbf{3} & Generated medical reports closely resemble actual reports in critical elements, albeit with minor deviations that do not compromise the overall accuracy and intelligibility. They maintain alignment with the principal diagnosis and recommendations of the authentic reports, despite possible negligible discrepancies in non-essential information. These reports encompass nearly all vital information, with minimal omissions that do not impede comprehensive understanding. & This criterion applies when the generated report's overall quality approximates that of the actual report, albeit with slight variances in detail. \bigstrut\\
    \hline
    \textbf{2} & The generated report diverges significantly from the actual report in several key aspects, yet its primary content and intent remain recognizable. Despite discrepancies in crucial information, the core diagnosis and recommendations are accurate. Omissions of some information elements have a marginal impact on the report's overall comprehensibility. & This level is relevant when the report's foundational structure mirrors that of the actual report, despite a notable quantity and severity of inaccuracies or omissions. \bigstrut\\
    \hline
    \textbf{1} & The generated report is quite different from the actual report in several key aspects, which affects the basic understanding and use of the reports. There are multiple errors or misleading statements of key information. More important information is left out, affecting the completeness and practicality of the report. & Although the report retains some basic framework or content, there are many errors, which affect the overall quality. \bigstrut\\
    \hline
    \textbf{0} & The generated report hardly has any consistency with the actual report and completely deviates from the correct message or purpose. The information in the report is completely inconsistent with the actual report and is full of errors or irrelevant content. It lacks the necessary information elements to serve as an effective medical report. & It is applicable when the content of the report is completely inconsistent with the intended goal and it is almost impossible to identify it as a valid medical report. \bigstrut\\
    \Xhline{1pt}
    \end{tabular}%
  \label{tab:addlabel}%
\end{table*}%

\end{document}